# Probabilistic Generative Deep Learning for Molecular Design

Daniel T. Chang (张遵)

*IBM (Retired)* dtchang43@gmail.com

**Abstract:**

Probabilistic generative deep learning for molecular design involves the discovery and design of new molecules and analysis of their structure, properties and activities by probabilistic generative models using the deep learning approach. It leverages the existing huge databases and publications of experimental results, and quantum-mechanical calculations, to learn and explore molecular structure, properties and activities. We discuss the major components of probabilistic generative deep learning for molecular design, which include molecular structure, molecular representations, deep generative models, molecular latent representations and latent space, molecular structure-property and structure-activity relationships, molecular similarity and molecular design. We highlight significant recent work using or applicable to this new approach.

## 1 Introduction

*Probabilistic generative deep learning for molecular design*, or *probabilistic generative molecular design (PGMD)* for short, involves the discovery and design of new molecules and analysis of their structure, properties and activities by probabilistic generative models using the deep learning approach. *Probabilistic generative deep learning* [1-2] has shown great promise for molecular design [4-5]. It can leverage the existing huge databases and publications of experimental results, and quantum-mechanical calculations, to learn and explore molecular structure, properties and activities.

The goal of probabilistic generative deep learning is to learn *generative concept representation*s [1-3], which are latent representations, using *deep generative models (DGMs)* [1]. A key characteristic of generative concept representations is that they can be directly manipulated to generate new concepts with desired attributes. There are two widely used DGM architectures: variational autoencoders and generative adversarial networks.

The *variational autoencoder (VAE)* [1] consists of an *encoder network* and a *decoder network* which encodes a data example to *latent representations* and generates samples from the latent space, respectively. The decoder network is a differentiable generator network and the encoder network is an auxiliary inference network. Both networks are jointly trained using *variational learning*, which is mainly applied to *prescribed DGMs*. Latent representations learned using VAEs generally are explicit, continuous and with meaningful structure. As such, they are suited for use as generative concept representations which can be directly manipulated to generate new concepts with desired attributes. *We focus on VAEs.*

The *generative adversarial network (GAN)* [1] consists of a differentiable *generator network* and an auxiliary *discriminator network* which generates samples from *latent representations* and discriminates between data samples and generated samples, respectively. Both networks are jointly trained using *adversarial learning*, which is mainly applied to *implicit DGMs*. Latent representations learned using GANs generally are implicit, discrete and without meaningful structure. As such, they may not be suited for use as generative concept representations.

In this paper we discuss the major components of *PGMD*, which include molecular structure, molecular representations, deep generative models, molecular latent representations and latent space, molecular structure-property and structure-activity relationships, molecular similarity and molecular design. We highlight significant recent work using or applicable to *PGMD*.

## 2 Background on Molecular Design

The following diagram shows the major components, and their interdependencies, of molecular design:

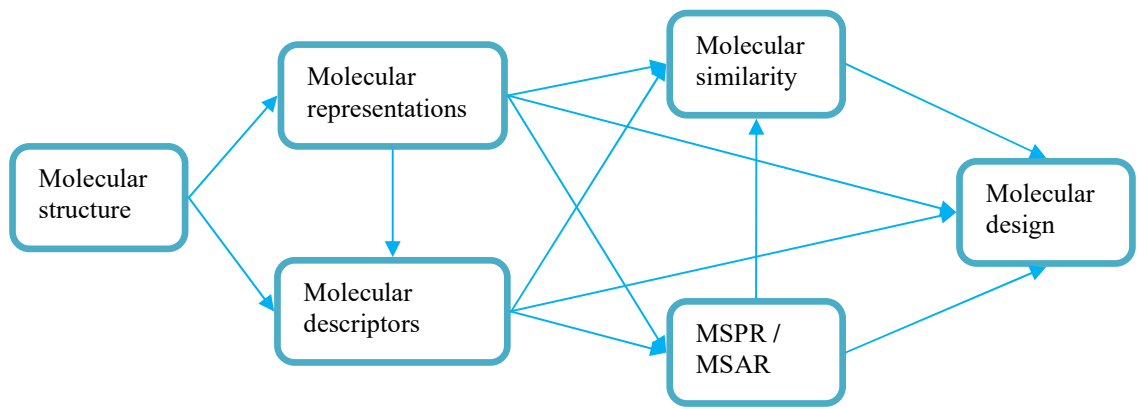

where MSPR stands for molecular structure-property relationships and MSAR for molecular structure-activity relationships.

### 2.1 Molecular Structure

*Molecules* are microscopic, material bodies with more or less well-defined structure. From a statistical perspective, we can define *molecular structure* [6] as that which distinguishes a molecule from a collection of its *constituent atoms*. Mathematically, molecular structure is measured by the inter-atom probability distribution. Thus an ideal gas of atoms has minimal structure; a hydrogen-bonded liquid is more structured; and a crystal or molecular solid is even more so.

*Electronic structure* and molecular structure are intimately related. The attractive forces exerted by the nuclei on electrons impart structure to the electron distribution in a molecule. The electron density $\rho(\mathbf{r})$ uniquely determines the ground state wave function $\psi$, the ground state electronic energy and the molecular structure, and vice versa.

The molecular structure is determined by three elements: constitution, configuration, and conformation [7]. *Constitution* means a certain manner and sequence of *chemical bonding* of atoms and is expressed by topological descriptors, presence and absence of functional groups / fragments, or other descriptors which account for the 2D features of a molecule. *Configuration* is defined by a 3D spatial arrangement of atoms, which is in turn characterized by the *valence angles* of all atoms that are directly linked to at least two other atoms. Configuration is expressed by shape descriptors. Finally, the *conformations* of a given molecule represent various thermodynamically stable 3D spatial arrangements of its atoms. For simplicity, *we focus on constitution* since it has the greatest impact on molecular properties and activities.

*Functional Groups*

Functional groups are a key aspect of molecular structure. A *functional group* is a collection of atoms at a site within a molecule with a *common bonding pattern*. An example is arenes which have alternating single and double C-C bonds in a six-member aromatic ring. The functional group reacts in a typical way, generally independent of the rest of the molecule. Functional groups give the molecule its *properties*, regardless of what molecule contains it; they are centers of *chemical reactivities*.

## 2.2 Molecular Representations

*Molecular representations* provide *machine-readable* representations of molecular structure. A given molecular structure can have many valid and unambiguous molecular representations. Molecular representations are *crucial to PGMD* since they are used for representing sample molecular structures as training data.

The most widely used molecular representations are line notations and molecular graphs. *Line notations* represent molecular structure as a linear string of characters; *molecular graphs* represent molecular structure as a graph. Line notations include the *Simplified Molecular-Input Line-Entry System (SMILES)* and the *IUPAC Chemical Identifier (InChI)*. Between the two, *we focus on SMILES* since they are widely used in deep learning.

## SMILES

*SMILES* [8] represents a *valence bond model* of molecule structure, which has proved to be an incredibly useful model for chemistry. A valence bond model is a way of allocating a molecule's nuclei and electrons into *atoms and bonds* in a way that makes sense. The function of SMILES is to clearly represent a particular valence bond model, not dictate which one should be used. SMILES includes specifications of the following elements / aspects of molecular structure: atoms, bonds, branching, rings, disconnections, isomerism and reactions. SMILES can be used in a great variety of ways: a character string, a list of tokens, a parse tree, or a molecular graph. The open-standards version of SMILES is defined by *OpenSMILES* (http://www.opensmil es.org/), which includes a brief description of *Canonical SMILES*. A molecular structure expressed in Canonical SMILES will always yield the same SMILES string.

A common approach used in *PGMD* is to train a generative model on SMILES and then use it to generate SMILES for new molecules with a desired property or activity. However, there is no guarantee that the resulting SMILES will be valid or even if they are, that they will represent a reasonable molecular structure. The *validity problems* can be broken down into those related to the *semantics* of SMILES, versus those that involve *invalid syntax*.

*DeepSMILES* [9] addresses two of the main causes for invalid syntax. Its *syntax* avoids the problem of unbalanced parentheses by only using close parentheses, where the number of parentheses indicates the branch length. In addition, it avoids the problem of pairing ring closure symbols by using only a single symbol at the ring closing location, where the symbol indicates the ring size. DeepSMILES can be converted to / from SMILES.

The *extraction of molecular structures* from publicly available documents such as journal articles and patent filings is an important area of data curation. Unfortunately, most publications do not provide the molecular structures in a machine-readable format. Instead, they contain hand-drawn molecular structure images. *Deep learning* has been used to provide solutions [17] for both *segmenting* molecular structure images from documents and for *predicting* molecular structures from these segmented images. The method takes an image or PDF and performs segmentation using a convolutional neural network (CNN). *SMILES* are then generated using a CNN in combination with a recurrent neural network (RNN) (encoder-decoder) in an end-to-end fashion.

*Molecular Graph*

The *molecular graph* [12-13] encodes molecular structure by a graph $G = (V, E, \mu, \upsilon)$ where the set of nodes V encodes the set of *atoms* and the set of edges E encodes the set of *bonds*. The labeling function $\mu(v)$ associates *atom features* (e.g., atom type) to each atom $v \in V$ and the labeling function $\upsilon(e)$ associates to each edge $e \in E$ the corresponding *bond features* (e.g., bond type).

*Molecular graph convolutions* [13] are a deep learning architecture for learning from molecular graphs, specifically for small molecules. They use a simple encoding of the molecular graph and extract meaningful molecule-level features to form *molecular representations* that can be used in deep learning applications, such as *PGMD*. The first basic unit of representation is an *atom layer*. The next basic unit of representation is an *atom pair (bond) layer*. The initial *atom features* include atom type, hybridization, hydrogen bonding, aromaticity, ring sizes, formal charge, partial charge and chirality. The initial *atom pair (bond) features* include bond type, graph distance and same ring. The *molecule-level features* are constructed only from the top-level atom features, after one final convolution on the atoms.

## 2.3 Molecular Descriptors

*Molecular descriptors* [14] are *numbers* that capture particular features of molecular structure. *Topological descriptors*, which are constitutional molecular descriptors, capture 2D features of molecular structure. Numerous topological descriptors have been proposed, including molecular fingerprints and molecular-graph based descriptors.

*Molecular Fingerprints*

*Molecular fingerprints* [7] stand for the presence or absence of some features (e.g. fragments) within a molecule. 2D or 3D features are encoded by setting bits (features) in a *bit-string (fingerprint)*. Molecular fingerprints are the most widely-used topological descriptors.

*Extended-connectivity fingerprints (ECFPs)* [15] are topological fingerprints designed for molecular characterization, similarity searching, and structure-activity modeling. They are among the most popular molecular fingerprints. ECFPs are based upon a process derived from the *Morgan algorithm*, one of the original methods for molecular comparison. They have many useful qualities: they can be rapidly calculated; they can represent a very large number of different features (up to 4 billion); features are not predefined, and so can represent variation in new structures; features can represent stereo-chemical information; and different initial atom identifiers can be used to generate different fingerprints, with different uses.

*Molecular Graph Based Descriptors*

Molecular descriptors can be generated from molecular graphs through construction of matrices and graph enumeration. Various *properties of the molecular graph*, such as degree counts for nodes, connectivity, atom types, etc., can be used as descriptors. These are referred to as *topological indices (TIs)* [29]. TIs can take many forms and are generally defined as some function of the nodes and edges in a molecular graph. Some of the well-known TIs include the Wiener index, Randic connectivity index, Kier higher-order connectivity indices, and shape index.

A *graph convolutional network (GCN)* [30] is used to generate differentiable *neural molecular descriptors* [16] based on molecular graphs. The architecture generalizes standard molecular feature extraction methods based on molecular fingerprints, with every non-differentiable operation replaced with a differentiable analog. Each feature of a neural molecular descriptor can be activated by similar but distinct *molecular fragments*, making the feature representation more meaningful than standard molecular fingerprints. Furthermore, the network allows end-to-end learning of *molecular predictors* whose inputs are molecular graphs. This network is of interest for their potential adaption for use in *PGMD*.

## 2.4 Molecular Structure-Property Relationships

The *molecular properties* of interest include physical properties and chemical properties. *Physical properties* are any properties of a molecule which can be observed and measured without changing its molecular structure. In contrast, *chemical properties* are those that can only be observed and measured by performing a *chemical reaction*, thus changing the structure of the molecule.

*Molecular structure-property relationships* [23] are indispensible to molecular similarity and molecular design which depend on target molecular properties. Chemists can identify *functional groups* or *fragments* related to molecular properties. Therefore, it is critical to correctly identify functional groups or fragments, which determine target molecular properties, to learn more accurate molecular structure-property relationships. Molecular structure-property relationships are commonly known as *QSPRs (Quantitative Structure-Property Relationships)*.

The most common types of QSPRs used in molecular design are *group contribution (GC)* methods [29]. These work under the assumption that a molecule's properties can be predicted by the number of occurrences of various molecular substructures called *(functional) groups*. GC methods define the *number of occurrences* $n_g$ of each of the groups g, which would also be associated with a *coefficient* $c_g$ that quantifies its "contribution" to a particular property P. Properties are calculated as: $P = \sum_g c_g n_g$. GC methods represent molecular structure in terms of its functional groups, very analogous to how chemists compare and analyze molecular structure. Different GC methods to estimate different properties usually have different sets of groups.

*Molecular graphs* have been used to produce a large number of QSPRs. In particular, *topological indices (TIs)* are paired with *regression coefficients* [29] and used to estimate properties in a similar way to GC methods. TIs can discriminate between very similar molecular structures, often in cases where GC methods cannot. However, they are not as generally applicable as GC methods and many of the graph properties are not always readily understandable from a chemical perspective.

### *Deep Learning Approaches*

The rise of *deep learning* offers a new viable *solution* to elucidate the molecular structure - property relationships directly from chemical data. The following discuss some relevant work. All are based on *molecular representations*, i.e., SMILES or molecular graphs. They are of interest because of their potential use in *PGMD*.

*Smiles2vec* [22] is a *RNN* that automatically learns features from *SMILES* strings to predict a broad range of molecular properties, including toxicity, activity, solubility and solvation energy.

A *GCN* is extended with the attention and gate mechanisms [23] to automatically extract features from *molecular graphs* related to a target molecular property such as solubility, polarity, synthetic accessibility and photovoltaic efficiency. The attention mechanism can differentiate atoms in different chemical environments by considering an interaction of each atom with neighbors. For example, the augmented GCN can recognize polar and nonpolar *functional groups* as important structural features for molecular solubility and polarity. As a result, it can accurately predict molecular properties and place molecular structures with similar properties close to each other in a well-trained *latent space*.

*GCNs* are used for discovering *functional groups* in organic molecules that contribute to specific *molecular properties* [31]. Molecules are represented as *molecular graphs*. The GCNs are trained in a supervised way on experimentally-validated molecular training sets (BBBP, BACE, and TOX21) to predict specific molecular properties, e.g., toxicity. Upon learning a GCN, its activation patterns are analyzed to automatically identify functional groups using four different methods: gradient-based saliency maps, class activation mapping (CAM), gradient-weighted CAM, and excitation back-propagation.

Absorption, distribution, metabolism and excretion (ADME) studies are critical for drug discovery. *Chemi-Net* [24] is a deep neural network architecture for ADME property prediction. It features a *molecular GCN* combined with the *multi-task deep neural networks (MT-DNNs)* method to boost prediction accuracy. For Chemi-Net, input *SMILES* strings are first converted to graph structures using a molecular conformation generator. The resultant *molecular graphs* are then used for training and testing.

## 2.5 Molecular Structure-Activity Relationships

The molecular activities of interest include chemical reactivity and biological activity. *Chemical reactivity* is the impetus for which a molecule undergoes a *chemical reaction*, either by itself or with other materials. *Biological activity* is the inherent capacity of a molecule, such as a drug or toxin, to alter one or more *chemical or physiological functions* of a cell, tissue, organ, or organism. The biological activity of a molecule is determined not only by the molecule's physical and chemical properties but also by its concentration and the duration of cellular exposure to it. Biological activity is driven by chemical reactivity.

*Molecular structure-activity relationships* are indispensible to molecular similarity and molecular design which depend on target molecular activities. Again, chemists can identify *functional groups* or *fragments* related to molecular activities. Therefore, it is critical to correctly identify functional groups or fragments, which determine target activities, to learn more accurate molecular structure-activity relationships. Molecular structure-activity relationships are commonly known as *QSARs (Quantitative Structure-Activity Relationships)* [25]. QSARs have typically been used for drug discovery and development in the *biological domain*.

For simplicity, we do not discuss further molecular structure-activity relationships. However, many of the approaches and solutions involving molecular structure-property relationships apply equally well to molecular structure-activity relationships, which are broader and more complex.

## 2.6 Molecular Similarity

The objective of *molecular similarity* [7] measures is to allow assessment of *molecular properties or activities*. The ideal measures, therefore, should be relevant to the molecular property or activity of interest. One of the basic beliefs of chemistry is that similarity in *molecular structure* implies similarity in molecular properties or activities. Thus, molecular similarity measures have been developed based on similarity between *molecular representations* or *molecular descriptors*. All these measures attempt to describe molecular structures by a set of *numerical values* and define some means for comparison between them.

*Molecular Representation Based*

With SMILES strings, the similarity between molecular structures can be computed using *SMILES-based string similarity functions* [26]. Various SMILES-based similarity methods have been adapted and evaluated for drug-target interaction prediction. Among these are the *cosine similarity based SMILES kernels* which obtain better scores than widely used 2D-based similarity kernels.

*Graph kernels* are a promising approach for tackling the *similarity and learning* tasks at the same time for *molecular graphs*. A general framework for designing graph kernels [27] utilizes the well-known *message passing scheme* on graphs. The kernels consist of two components. The first component is a kernel between vertices, while the second component is a kernel between graphs. The framework is evaluated on the QM9 dataset for *molecular graph prediction*. Each molecule in the dataset has 13 target properties to predict. The framework achieves lower mean absolute error and root mean squared error values than all the baselines on 10 out of the 13 targets. This framework is of interest and has potential use in *PGMD*.

*Molecular Descriptor Based*

Constitutional molecular similarity assessment is based on *topological descriptors*. When *molecular fingerprints* are used, molecules are estimated to be structurally similar if they have many bits in the fingerprints in common. Fingerprints are usually compared by the *Tanimoto coefficient* [7]:

$$\tau = \frac{N_{A\&B}}{N_A + N_B - N_{A\&B}}$$

where $N_A$ is the number of features (bits) in the fingerprint A, $N_B$ is the number of features in B, and $N_{A\&B}$ is the number of features common to A and B.

## 2.7 Molecular Design

The objective of *molecular design* is to discover and design new molecules with desired *molecular properties or activities*. The task is challenging since the *chemical space* is vast and often difficult to navigate. One of the basic beliefs of chemistry is that *molecular structure* largely determines molecular properties or activities. Therefore, molecular design is based on *molecular representations* or *molecular descriptors* and optimized for target *molecular structure-property relationships* or *molecular structure-activity relationships*.

Our focus is on *PGMD*, which is based on molecular representations, as discussed in the rest of the paper.

# 3 PGMD Components and Architecture

The major components, and their interdependencies, of *PGMD* are shown below:

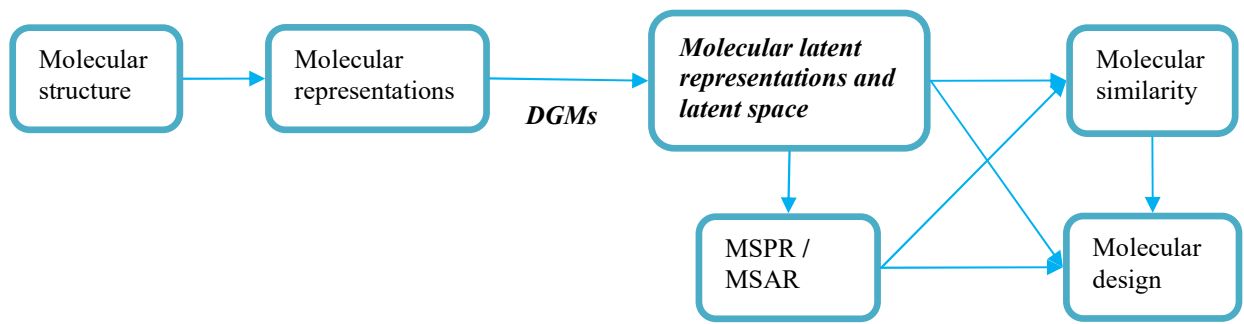

*DGMs* and *molecular latent representations and latent space* are new and unique to *PGMD*. Molecular latent representations and latent space make it unnecessary and inappropriate to use molecular descriptors in *PGMD*.

## 3.1 DGM Architecture for PGMD

The *DGM architecture for PGMD* is shown below [5]:

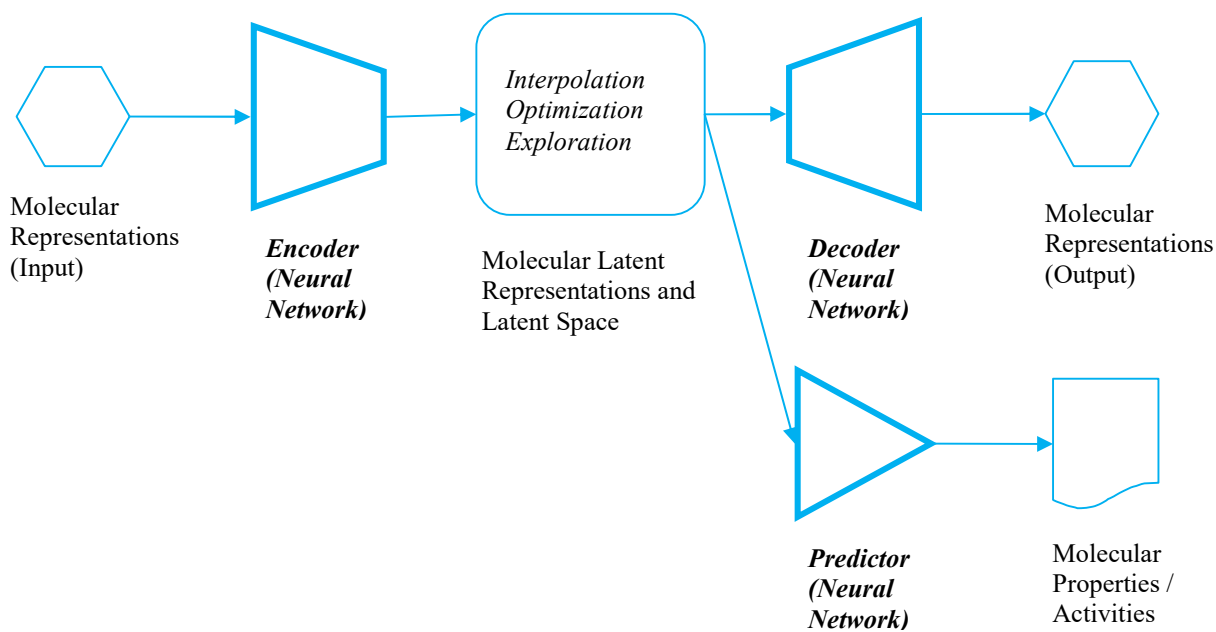

The DGM architecture extends the standard *VAE architecture* (*encoder – decoder*) with the *predictor for molecular properties / activities*, with all three components jointly trained.

See [1-2] for discussions of various *VAEs* and their pros and cons, and [2] for discussions of various *latent variable models (LVMs)* used in VAEs and their pros and cons.

## 3.2 Molecular Latent Representations and Latent Space

The nature and characteristics of latent representations and latent space learned in a DGM are determined by the LVM used in the DGM, the sample data representations, and the DGM architecture and algorithms. See [2] for discussions, including *disentangled and hierarchical latent representations, latent space interpolation, latent space vectors* and *latent space geometry*, as well as their significance to *generative concept representations*.

Molecular latent representations and latent space are the center piece of *PGMD*. *Molecular latent space* supports interpolation, optimization and exploration, and is the *foundation* of molecular structure-property and structure-activity relationships, molecular similarity and molecular design.

# 4 PGMD: SMILES Based

SMILES is the most widely used molecular representation. For representing sample molecular structures as training data, SMILES can be used in a great variety of ways: a character string, a list of tokens, or a parse tree.

## 4.1 CVAE

SMILES strings are *sequences* of characters. The *CVAE* [1, 5, 18] adapts the VAE to sequence by using single-layer Long-Short Term Memory (LSTM) RNNs for both the encoder and the decoder. Latent representations learned using the CVAE contain entire sequences and can be used to generate coherent new sequences that interpolate between known sequences.

Continuous *molecular latent spaces* are learned using the CVAE: one with 130K molecules from the QM9 dataset of molecules with fewer than 9 heavy atoms and another with 250K drug-like molecules extracted at random from the ZINC dataset. Interpolating linearly between two latent points is found to be inadequate. Instead, *spherical linear interpolation (slerp)* [2] is used, which treats the interpolation as a circle path on an *n-dimensional hypersphere*. The CVAE is trained jointly on the reconstruction task and an additional molecular-property prediction task. With joint training for property prediction, the distribution of molecules in the molecular latent space is *organized by molecular property values*.

The CVAE is trained on molecular structures (as SMILES strings) from the QM9 and ZINC datasets by using an encoder, a decoder, and a predictor. The *encoder* converts the SMILES string into a continuous *molecular latent representation*, and the *decoder* converts these molecular latent representations back to SMILES strings. The *predictor* estimates *molecular properties* from the molecular latent representations. The continuous *molecular latent space* allows one to automatically *generate new molecular structures* by performing simple latent space operations, such as decoding random latent space points, interpolating between latent space points, or perturbing known latent space vectors. To compensate for the limitation of SMILES strings, the RDKit is used to *validate* the output molecular structures and invalid ones are discarded.

The continuous molecular latent space also allows the use of powerful *gradient-based Bayesian optimization* to efficiently guide the search for optimized molecular structures with *desired molecular properties*. In order to create a smoother landscape to perform optimizations, a *Gaussian process model* is used to model the property predictor model. The Gaussian process is trained to predict *target properties* for molecules given the *molecular latent representation* as an input.

## 4.2 GVAE

SMILES is *structured data*, not pure sequences. As a result, the CVAE will often lead to invalid outputs because of the lack of formalization of syntax and semantics serving as constraints. *Context-free grammars* can be used to incorporate syntax constraints. To do so, SMILES is represented as *parse trees* from the grammar. The *GVAE* [1, 10] encodes and decodes directly from and to these parse trees, respectively, ensuring the generated outputs are always *valid based on the grammar*.

By ensuring the generated outputs are always *syntactically valid*, the GVAE learns a more coherent *molecular latent space* than the CVAE, in which nearby points decode to similar discrete outputs. This molecular latent space is very *smooth*; in many cases moving from one latent point to another will only change a single atom in a molecule. The training data are 250K drug-like molecules extracted at random from the ZINC dataset.

The GVAE uses the same methodology used by the CVAE, and the ZINC dataset, for molecular design. Due to the use of SMILES parse trees which ensure the generated outputs are always *syntactically valid*, the GVAE generates about twice more valid molecular structures than the CVAE and achieves *better optimization* by finding new molecular structures with better molecular properties.

## 4.3 SD-VAE

Although the GVAE provides the mechanism for generating syntactically valid outputs, it is incapable to constraint the model for generating semantically valid outputs. *Attribute grammars* allow one to attach semantics to a parse tree generated by context-free grammar. The *SD-VAE* [1, 11] incorporates attribute grammar in the VAE such that it addresses both syntactic and semantic constraints and generates outputs that are *syntactically valid* and *semantically coherent*.

The SD-VAE makes further improvement than the GVAE as a result of generating outputs that are both syntactically valid and semantically coherent. The *molecular latent space* learned is *smoother and more discriminative*. The training data are the same as in the GVAE.

The SD-VAE follows the protocols used by the GVAE for molecular design. Due to the use of SMILES annotated parse trees which ensures the generated outputs are both *syntactically valid* and *semantically coherent*, the SD-VAE finds *even better solution* in molecular optimization and prediction tasks.

# 5 PGMD: Molecular Graph Based

SMILES-based molecular design has two critical limitations [19]. First, SMILES is not designed to capture *molecular similarity*. This prevents VAEs from learning smooth molecular embeddings. Second, *molecule validity* is easier to express on molecular graphs than linear SMILES. Therefore, operating directly on *molecular graphs* improves *generating valid molecular structures*.

## 5.1 VAE with Regularization Framework

Generating semantically valid graphs is a challenging task for DGMs. A *regularization framework* [20] is used for training VAEs with *GCNs* that encourages the satisfaction of *validity constraints*. The approach is motivated by the transformation of a constrained optimization problem to a regularized unconstrained one. It focuses on the *matrix representation* of graphs and formulates *penalty terms* that regularize the output distribution of the decoder to encourage the satisfaction of validity constraints. The penalties in effect regularize the distributions of the *existence and types of the nodes and edges* collectively. Examples of validity constraints include *graph connectivity* and *valence* in the context of molecular graphs.

## 5.2 CGVAE

The *CGVAE* [21] is a *sequential* generative model for molecular graphs built from a VAE with *gated graph neural networks (GGNNs)* [30] in the encoder and decoder. It learns latent representations of attributed *nodes (atoms)* instead of entire molecules. The decoder forms nodes and edges alternately. *Decoding* is performed by first initializing a set of possible nodes to connect. The decoder then iterates over the given nodes, performs a step of edge selection and edge labeling for the currently focused node, passes the current connected molecular graph to a GGNN for updating the node representations, and repeats this process until an edge to a special stop node is selected. This entire process is repeated for a new node in the current connected graph and terminates if there are no valid candidates. To help ensure valid molecule generation the decoder makes use of a *valency mask* to prevent generation of additional bonds on atoms that have already been assigned the maximum number of bonds for that particular atom type.

Continuous *latent spaces* are learned using the CGVAE from the QM9, ZINC and CEPDB (a subset of the database containing 250K randomly sampled organic molecules) datasets. The latent representations consist of attributed *atoms (nodes)* instead of entire molecules. The CGVAE is trained jointly on the reconstruction task and an additional molecular-property prediction task. With joint training for property prediction, the distribution of atoms in the latent space is *shaped by* *molecular properties*. The *major drawback* of the atom-embedded latent space is that one cannot perform molecule-level latent space interpolation and arithmetic on the latent space.

The CGVAE is trained on molecular structures (as molecular graphs) from the QM9, ZINC and CEPDB datasets by using an encoder, a decoder, and a predictor. The *encoder* converts a molecular graph into a continuous *latent representation*, and the *decoder* converts these latent representations back to molecular graphs. The *predictor* estimates *molecular properties* from the latent representations. The continuous *latent space* allows the use of the *gradient-based GGNN regression model* [28] to optimize the latent space and direct the generation towards especially interesting molecular structures. The CGVAE is excellent at matching graph statistics, while generating *valid, novel and unique* molecular structures for all datasets considered.

### 5.3 JT-VAE

The *JT-VAE* [19] generates molecular graphs in two phases. First, it generates a tree-structured object (a *junction tree*) that represents the scaffold of *subgraph components* and their coarse relative arrangements. The components are *valid chemical substructures* automatically extracted from the training set using tree decomposition. Second, the subgraphs (nodes in the junction tree) are assembled into a coherent *molecule graph* using a graph message passing network. This approach incrementally generates the molecular graph while maintaining *molecular validity* at every step. The subgraph components are used as *building blocks* both when *encoding* a molecular graph into a latent representation as well as when *decoding* latent representations back into valid molecular graphs. Thus, the JT-VAE encoder has two parts: *graph encoder* and *tree encoder*, so has the decoder: *tree decoder* and *graph decoder*. The graph and tree encoders are closely related to *message passing neural networks (MPNNs)* [28, 30].

The *JT-VAE* is used to learn a continuous *molecular latent space* from the ZINC dataset. In JT-VAE, both the *molecular graph* and its associated *junction tree* offer two complementary representations of molecular structure. Therefore the molecular structure is encoded into a *two-part continuous latent representation $z = [z_T, z_G]$* where $\mathbf{z}_T$ encodes the tree structure and what the subgraph components are in the tree. $\mathbf{z}_G$ encodes the graph to capture the fine-grained connectivity. The molecular latent space learned is *smoother* than that of SD-VAE.

The JT-VAE decomposes the challenge of molecular design into two complementary subtasks: learning to represent *molecular graphs* in a *continuous manner* that facilitates the prediction and optimization of their properties (*encoding*); and learning to map an *optimized continuous representation* back into a *molecular graph* with *desired properties* (*decoding*). The JT-VAE generates *100% valid molecular structures* when sampled from a prior distribution, outperforming the baselines (CVAE, GVAE and SD-VAE) by a significant margin for the ZINC dataset.

Similar to the baselines, the continuous molecular latent space allows the use of powerful *gradient-based Bayesian optimization* to efficiently guide the search for optimized molecular structures. A *Gaussian process* is trained to predict target properties for molecules given the molecular latent representation as an input. The JT-VAE excels in discovering *new molecular structures with desired properties*, yielding a 30% relative gain over the baselines for the ZINC dataset.

## 6 Summary and Conclusion

*PGMD* involves the discovery and design of new molecules and analysis of their structure, properties and activities by probabilistic generative models using the deep learning approach. It leverages the existing huge databases and publications of experimental results, and quantum-mechanical calculations, to learn and explore molecular structure, properties and activities. In this paper we discussed the major components of *PGMD* and highlighted significant recent work using or applicable to *PGMD*.

*PGMD* is a new and promising approach to molecular design. Two areas deserve significant future work and exploration. The first is molecular graphs due to their flexibility and power for representing molecular structure and the rapid advances in graph neural networks. The second is molecular latent representations and latent space which are central to *PGMD* and are the foundation of molecular structure-property and structure-activity relationships, molecular similarity and molecular design

There are other areas awaiting future work and exploration as well. These include hierarchical molecular latent representations (e.g., functional groups), molecular latent space vectors (e.g., molecule vector, group vector, attribute vector), molecular latent space geometry (e.g., geodesic interpolation), and molecular similarity measures based on molecular latent representations and latent space.